\documentclass{article} 
\usepackage{iclr2024_conference,times}

\usepackage{amsmath,amsfonts,bm}

\def\eqref#1{equation~\ref{#1}}

\def\1{\bm{1}}

\DeclareMathAlphabet{\mathsfit}{\encodingdefault}{\sfdefault}{m}{sl}
\SetMathAlphabet{\mathsfit}{bold}{\encodingdefault}{\sfdefault}{bx}{n}

\usepackage{microtype}
\usepackage{graphicx}
\usepackage{mwe}
\usepackage{subcaption}
\usepackage{booktabs} 
\usepackage{adjustbox}

\usepackage{suffix}
\usepackage{ragged2e}
\usepackage[linesnumbered,ruled,vlined]{algorithm2e}
\SetKwComment{tcc}{\scriptsize /* }{\tiny */}%
\SetKwComment{tcp}{\scriptsize // }{}%
\newcommand{\stcp}[1]{\tcp{\scriptsize #1}}
\WithSuffix\newcommand\stcp*[1]{\tcp*{\scriptsize #1}}

\usepackage[utf8]{inputenc} 
\usepackage[T1]{fontenc}    
\usepackage[pagebackref,breaklinks,colorlinks]{hyperref}

\definecolor{fig_blue}{HTML}{0057AD}
\definecolor{fig_green}{HTML}{668C50}
\definecolor{fig_beige}{HTML}{D4BFA9}
\definecolor{fig_gray}{HTML}{8A8A8A}

\usepackage{graphicx}
\usepackage{amsmath}
\usepackage{amssymb}
\usepackage{booktabs}
\usepackage{bbm}
\usepackage[utf8]{inputenc} 

\usepackage{adjustbox}

\usepackage[nameinlink]{cleveref}
\Crefname{section}{Section}{Sections}

\crefname{table}{Table}{Tables}
\Crefname{figure}{Figure}{Figures}

\usepackage{pifont}

\usepackage{url}

\usepackage{tikz}
\usepackage{comment}
\usepackage{amsfonts}       
\usepackage{color}

\usepackage{amsmath}
\usepackage{amssymb}
\usepackage{bbm, dsfont}
\usepackage{mathtools}
\usepackage{amsthm}
\usepackage[normalem]{ulem}
\usepackage{xcolor,colortbl}
\usepackage{multirow}
\usepackage{cutwin}
\usepackage{arydshln}
\usepackage{caption}
\usepackage{enumitem}
\usepackage{wrapfig}
\usepackage{microtype}
\usepackage{graphicx}
\usepackage{bm}
\usepackage{eqparbox}
\usepackage{makecell}
\usepackage{threeparttable}

\newcommand{\eat}[1]{}

\definecolor{Red}{rgb}{0.6,0,0}
\definecolor{Blue}{rgb}{0,0,0.8}
\definecolor{Green}{rgb}{0.2,0.8,0}
\definecolor{airforceblue}{rgb}{0.36, 0.54, 0.66}
\definecolor{ao(english)}{rgb}{0.0, 0.5, 0.0}
\definecolor{azure(colorwheel)}{rgb}{0.0, 0.5, 1.0}
\definecolor{crimson}{rgb}{0.86, 0.08, 0.24}
\definecolor{darkcerulean}{rgb}{0.03, 0.27, 0.49}
\definecolor{cobalt}{rgb}{0.0, 0.28, 0.67}
\definecolor{rosegold}{rgb}{0.72, 0.43, 0.47}
\definecolor{orange-red}{rgb}{1.0, 0.27, 0.0}
\definecolor{mountainmeadow}{rgb}{0.19, 0.73, 0.56}
\definecolor{malachite}{rgb}{0.04, 0.85, 0.32}
\definecolor{darkblue}{rgb}{0.0, 0.0, 0.55}
\definecolor{customblue}{rgb}{0.2, 0.35, 0.8}
\definecolor{gg}{gray}{0.92}
\newcolumntype{a}{>{\columncolor{gg}}c}

\definecolor{gg}{HTML}{E4F1EE}
\definecolor{gg}{HTML}{FFF1F5}
\definecolor{ggg}{gray}{0.65}
\newcolumntype{a}{>{\columncolor{gg}}c}

\hypersetup{colorlinks=true}
\hypersetup{linktoc=all}
\hypersetup{citecolor=mountainmeadow}
\hypersetup{linkcolor=crimson}
\hypersetup{urlcolor=darkblue}
\usepackage[all]{hypcap}

\creflabelformat{equation}{#2\textup{#1}#3}  

\usepackage{svg}

\newcommand{\MethodName}{ECoFLaP}

\newcommand\blfootnote[1]{%
  \begingroup
  \renewcommand\thefootnote{}\footnote{#1}%
  \addtocounter{footnote}{-1}%
  \endgroup
}

\title{ECoFLaP: Efficient Coarse-to-Fine Layer-Wise Pruning for Vision-Language Models}

\iclrfinalcopy

\author{Yi-Lin Sung \;\;\;\;\; Jaehong Yoon \;\;\;\;\; Mohit Bansal\\
Department of Computer Science, UNC Chapel Hill\\
\texttt{\{ylsung, jhyoon, mbansal\}@cs.unc.edu} \\
}

\newcommand{\norm}[1]{\left\lVert#1\right\rVert}

\begin{document}

\maketitle

\begin{abstract}
Large Vision-Language Models (LVLMs) can understand the world comprehensively by integrating rich information from different modalities, achieving remarkable advancements on various multimodal downstream tasks.  However, deploying LVLMs is often problematic due to their massive computational/energy costs and carbon consumption. Such issues make it infeasible to adopt conventional \textit{iterative global pruning}, which is costly due to computing the Hessian matrix of the entire large model for sparsification. 
Alternatively, several studies have recently proposed layer-wise pruning approaches to avoid the expensive computation of global pruning and efficiently compress model weights according to their \emph{importance} within a layer. However, they often suffer from suboptimal model compression due to their lack of a global perspective. To address this limitation in recent efficient pruning methods for large models, we propose \textit{\textbf{E}fficient \textbf{Co}arse-to-\textbf{F}ine \textbf{La}yer-Wise \textbf{P}runing (\textbf{ECoFLaP})}, 
a two-stage \emph{coarse-to-fine} weight pruning approach for LVLMs. We first determine the sparsity ratios of different layers or blocks by leveraging the global importance score, which is efficiently computed based on the zeroth-order approximation of the global model gradients. Then, the model performs local layer-wise unstructured weight pruning based on globally-informed sparsity ratios. 
We validate our proposed method across various multimodal and unimodal models and datasets, demonstrating significant performance improvements over prevalent pruning techniques in the high-sparsity regime. \blfootnote{Our project page and code are available at 
\url{https://ecoflap.github.io/}}
\end{abstract}
\vspace{-0.4cm}
\section{Introduction} \label{sec: intro}
Deep learning models have been increasingly growing in size~\citep{radford2018improving,brown2020language,liu2023visual,zhu2023minigpt} in order to have a sufficient capacity to learn challenging tasks in the real world. However, this exorbitant model size requires significant computations and memory to deploy, which limits their applicability in many resource-constrained environments. Model compression, a strategy to reduce the size of neural networks while preserving their capabilities~\citep{dai2018compressing, park2022bitat, xiao2023smoothquant, fang2023depgraph}, has gained popularity in tackling this problem. In order to build lighter, faster, and more interpretable models, various directions have been studied in parallel, including model pruning~\citep{dai2018compressing,fang2023depgraph,frantar2023sparsegpt}, quantization~\citep{park2022bitat,xiao2023smoothquant}, layer drop~\citep{sajjad2023effect}, and token merging~\citep{bolya2022token,bolya2023token}. Among them, in this paper, we study an unstructured model pruning approach that has strong potential to preserve model performance, even with a high compression rate of large vision-language models.

While most model pruning approaches tackle the problem of vision-based tasks~\citep{liang2022not,kong2022spvit,wei2023joint} and a few recent works have studied to reduce the size of language models~\citep{ma2023llm,sun2023simple,frantar2023sparsegpt}, efficient pruning for multimodal models like Large Vision-Language Models (LVLMs)~\citep{ye2023mplug, li2023otter} have been understudied due to their architectural complexity and the disparities in data characteristics between different modalities~\citep{shi2023upop}. 
Unlike unimodal learning methods, 
network architectures for multimodal learning~\citep{li2022blip,Li2023BLIP2BL} are often composed of modality-specific sub-modules to capture the knowledge for each modality, which substantially expands the scale of multimodal models. Moreover, this modularization leads to significant imbalances in the weight/gradient distributions between modules associated with different modalities, making unified pruning difficult.

As conventional iterative pruning~\citep{mallya2018packnet,molchanov2019importance,wang2020picking,zhang2022advancing} demands extensive computations to learn pruning masks by computing the inverse Hessian in large models, layer-wise one-shot pruning~\citep{chen2018shallowing,frantar2023sparsegpt,sun2023simple} has recently been developed to compress them efficiently. These approaches discard uninformative model weights in a single iteration without computing the expensive second-order Hessian matrix for the entire model architecture, yet often suffer from finding the optimal sparsity ratio per layer, as they resort to intra-layer weight importance without sufficient understanding of global (i.e., inter-layer) weight correlations in the model, hence leading to suboptimal pruning.

To overcome this limitation of layer-wise pruning and build an efficient approach for LVLM compression, 
we propose \textit{\textbf{E}fficient \textbf{Co}arse-to-\textbf{F}ine \textbf{La}yer-Wise \textbf{P}runing (\textbf{\MethodName{}})} that obtains an adaptive sparsity ratio per layer in a single step based on a `global importance score' (\textit{Coarse}) and then removes parameters that are less critical (to the model's performance), in a layer-wise manner (\textit{Fine}). 
Note that our proposed method is computationally efficient by leveraging the first-order gradient to obtain a global importance score without Hessian operations, allowing layer-wise pruning to leverage a holistic correlation across the model weights.
To further improve the memory efficiency of the global score computation, we introduce the zeroth-order approximate gradient computed by the forward-forward algorithm~\citep{Malladi2023FineTuningLM, Hinton2022TheFA}. 
{In the end, our \MethodName{} obtains highly compressed large models in a single shot, with adaptive sparsity in each layer, leveraging global weight importance approximated by memory-efficient forward-forward operations. We note that our proposed pruning method generalizes well not only to vision-language multimodal tasks but also to unimodal tasks, including ImageNet (Vision) and MMLU (NLP).}

We extensively validate the efficacy of our approach across a variety of models and datasets, including but not limited to EVA-ViT, FlanT5, LLaMA, BLIP/BLIP-2, CLIP, and on image classification (ImageNet-1k), multitask multiple-choice question answering (MMLU), next token prediction (WikiText), visual reasoning (NLVR$^2$), visual question answering (VQAv2, OK-VQA, GQA), image captioning (NoCaps, COCO Captions), and image-text retrieval (Flickr30k). Our proposed \MethodName{} surpasses iterative global pruning and SoTA layer-wise pruning methods, \textit{SparseGPT}~\citep{frantar2023sparsegpt} and \textit{Wanda}~\citep{sun2023simple}, with relative improvements up to $5\%$ in average performance across multiple VL tasks.
Additionally, our approach outperforms a recent SoTA in pruning vision-language model, UPop~\citep{shi2023upop},
achieving relative improvements of 1.8\% and 2.6\% on NLVR$^{2}$ and COCO captions. \MethodName{} also generalizes well to unimodal models (FlanT5, LLaMA, EVA-ViT), particularly in a higher sparsity regime. It is worth noting that \MethodName{} with the zeroth-order gradient reduces the GPU memory usage by up to $40\%$ compared to baselines using the first-order gradient to estimate the global importance. These results not only underscore the potential of our method but also pave the way for more efficient and effective pruning techniques in the future.

We summarize our contributions as threefold:
\begin{itemize}[leftmargin=*]
\item We propose a novel layer-wise pruning method for vision-language models, coined \textbf{E}fficient \textbf{Co}arse-to-\textbf{F}ine \textbf{La}yer-Wise \textbf{P}runing for Vision-Language Models (\textbf{\MethodName{}}), that finds adaptive sparsity per layer by leveraging the global importance score approximated via zeroth-order gradients.
\item We validate the proposed method on various backbone structures with diverse vision-language and unimodal tasks, and demonstrate the superiority of our method by consistently outperforming the SoTA pruning baselines for large models on multiple benchmark datasets.
\item We provide extensive analyses and ablation studies that help to understand the challenges of unified multimodal model pruning and the strengths of our proposed methods compared to baselines, including visualizations of importance score distribution, dynamic sparsity, loss landscape, etc. 
\end{itemize}

\section{Related Work} \label{sec: related work}

\paragraph{Model pruning for transformers.} 

Efficient model inference via pruning~\citep{yoon2017combined, dai2018compressing, zhang2022platon, fang2023depgraph} is one of the long-standing topics in the community and is coming to the fore as transformer-based models grow in size. Model pruning is typically categorized into \textit{structured pruning} and \textit{unstructured pruning}. The former~\citep{mccarley2019structured,wang2019structured,kwon2022fast} aims to accelerate inference speed and throughput at the expense of model performance. Unstructured pruning~\citep{sun2023simple,frantar2023sparsegpt}, on the other hand, can be advantageous in preserving performance even with high model sparsity given AI acceleration software or sparse matrix computation schemes~\citep{han2016eie,mishra2021accelerating,das2022coded,deepsparse2022}. 
Recently, \cite{frantar2023sparsegpt} suggest a one-shot pruning technique, SparseGPT, for generated pre-trained transformers (GPTs) in an unstructured manner. They newly employ a sparse regression solver that prunes weights at each layer based on row-wise Hessian reconstruction as formulated by a closed-form solution. 
Wanda~\citep{sun2023simple} proposes a magnitude-based unstructured pruning approach for large language models (LLMs). It promotes layer-wise weight sparsification based on the importance, computed by multiplying weights and input activations. 
However, these SoTA pruning methods rely on the pre-defined pruning ratio that all layers resort to the same sparsity, restricting the upper bound on model compression. In addition, their methods are tailored to language models without concern for different modalities. On the other hand, our proposed method allows adaptive pruning at each layer without heavy computation of global gradients. Further, to the best of our knowledge, we propose a first unified sparse approximate solver for vision-language multimodal models.

\paragraph{Transformers for vision-language multimodal learning.}

Vision-language multimodal learning has shown remarkable achievement on various tasks, such as classification~\citep{liu2018learning,liang2022expanding}, retrieval~\citep{fei2021cross}, few-shot learning~\citep{tsimpoukelli2021multimodal,alayrac2022flamingo}, visual QA~\citep{kim2016multimodal,liu2023visual}, and image/video generation~\citep{zhou2022towards,singer2022make,lee2023text}. 
Recently, transformer modules have become the new standard for both language and visual-based~\citep{dosovitskiy2020image,liu2021swin} tasks, and the powerful representation of these transformers allows for the modularization of multimodal learning frameworks with existing pre-trained uni-modal models without an expensive re-training phase for multimodal data~\citep{radford2021learning,zhou2022learning}. 

\section{Background: Layer-wise Pruning for Vision-language Models} \label{sec: preliminaries}

\subsection{Preliminaries: A Framework for Vision-language Models} \label{ssec: framework of vl models}
Multiple vision-language multimodal models introduce different architectural designs to understand compositional multimodal semantics, which can affect the pruning strategy. 
In this paper, we basically build our pruning method upon two popular multimodal learning frameworks for vision-language tasks, BLIP~\citep{li2022blip} and BLIP-2~\citep{Li2023BLIP2BL}, which are modularized with the pre-trained ViT~\citep{dosovitskiy2020image} and FlanT5~\citep{Chung2022ScalingIL} architectures to encode visual and language information, respectively. They have achieved remarkable zero-shot performance on various vision-language tasks. Let the visual encoder $f_v(\cdot;\textbf{W}^v)$ and language model $f_l(\cdot;\textbf{W}^l)$ in BLIP variants be composed of a set of the weights $\textbf{W}^{v} = \{\textbf{W}^v_i | 1 \leq i \leq M\}$ and $\textbf{W}^{l} = \{\textbf{W}^l_i | 1 \leq i \leq N\}$, respectively, where $M$ and $N$ indicate the number of the corresponding layers. While freezing these pre-trained backbones, BLIPs introduce Query transformer (Q-Former) $f_q(\cdot;\textbf{W}^q)$, a lightweight module with a set of the weights $\textbf{W}^{q}$, that extracts essential information from visual representation and warps it to the input space of the language model~\citep{Li2021AlignBF}. Given the multimodal input pair $(\textbf{x}^v, \textbf{x}^l)$, the output of the model is represented as $f_l([\textbf{o}^v, \textbf{x}^l]; \textbf{W}^{l})$, where $\textbf{o}^v$ denotes the aligned visual representation $f_q(f_v(\textbf{x}^{v}; \textbf{W}^v); \textbf{W}^q)$. 

\subsection{Challenges in Pruning Vision-language Models} \label{ssec: layer-wise pruning}

Let us first describe a primary strategy for Hessian-based global pruning, which aims to remove the optimal subset\footnote{The subset size (i.e., sparsity) is often pre-defined or obtained through training.} of parameters minimizing the loss change $\delta \mathcal{L} = \mathcal{L} (\textbf{w} + \delta \textbf{w}) - \mathcal{L} (\textbf{w})$. Here, $\textbf{w}$ denotes the weight vector of the model. When the $i^{th}$ model parameter is removed, implying $\textbf{e}^{\intercal}_i \delta \textbf{w} + \textbf{\textbf{w}}_i = 0$, where $\textbf{e}^{\intercal}_i$ indicates the $i^{th}$ canonical basis vector, its importance is represented as ${\textbf{w}^2_i} / {(2 \cdot [\text{H}^{-1}]_{ii})}$. After that, the model prunes its weights with the lowest importance. However, computing the inverse Hessian $\text{H}^{-1}$ requires significant computational budgets and is often infeasible for large models. 

On the other hand, layer-wise one-shot pruning aims to leverage a small amount of calibration data (such as 128 data points) to remove less important weights from the model in a single step based on their local significance per layer. This approach is significantly efficient in terms of computation/memory in the model pruning phase without expensive iterations of the re-training or Hessian matrix computation for the parameters of the entire model, which is particularly beneficial for large language/multimodal models with billions of parameters.
The model compression procedure of recent layer-wise pruning methods~\citep{lazarevich2021post,yu2023x,frantar2023sparsegpt} can be summarized in the three steps given a target sparsity of $p\%$ and a layer index $i=1,~i\in\{1,...,L\}$: (1) Compute the importance score for each parameter within the layer $i$, (2) Discard $p\%$ of the parameters of layer $i$ based on their importance score and adjust the remaining weights (usually for Hessian-based methods), and (3) calculate the layer's output with only activated (i.e., not pruned) weights, and repeat the process for the next layer ($i\leftarrow i+1$) until $i=L$. 

\begin{figure}
     \centering
     \vspace{-0.05in}
     \begin{subfigure}{0.32\linewidth}
        \includegraphics[width=0.95\linewidth]{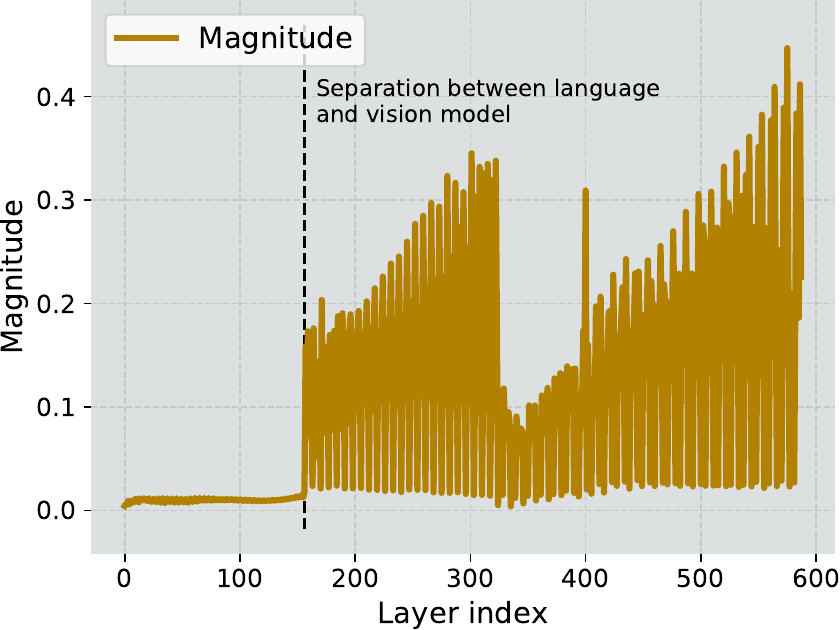}
        \caption{Weight Magnitudes}
        \label{fig: dist mag}
     \end{subfigure}
     \hfill
     \begin{subfigure}{0.32\linewidth}
        \includegraphics[width=\linewidth]{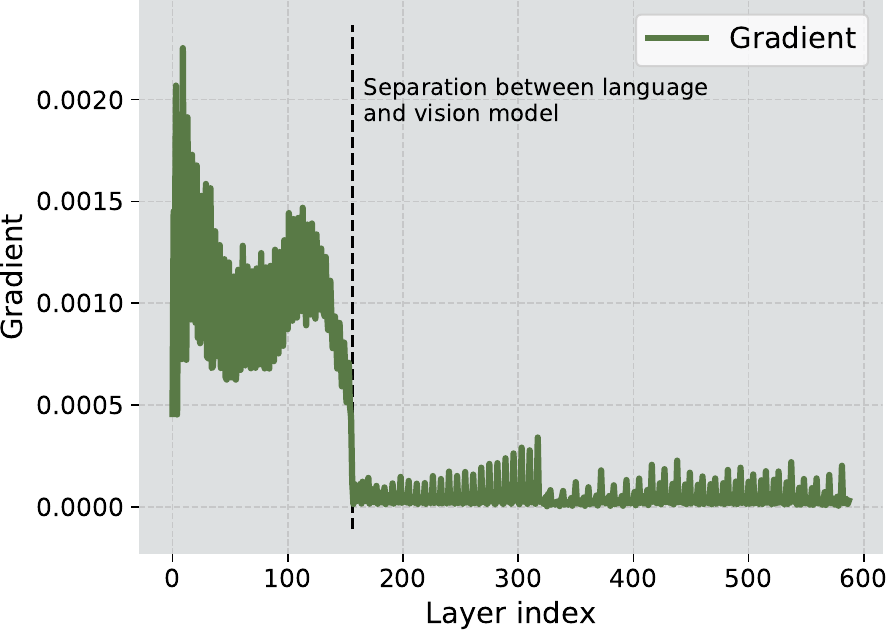}
        \caption{Weight Gradients}
        \label{fig: dist grad}
     \end{subfigure}
     \hfill
     \begin{subfigure}{0.32\linewidth}
        \includegraphics[width=0.962\linewidth]{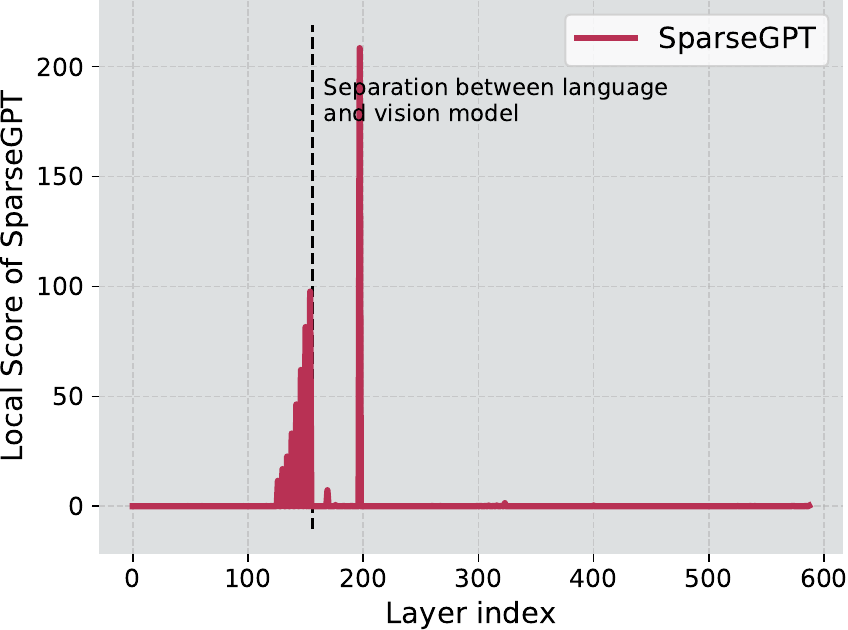}
        \caption{Local Score of SparseGPT}
        \label{fig: dist_sparsegpt}
     \end{subfigure}
     \vspace{-0.05in}
     \caption{\textbf{(a)} and \textbf{(b): The imbalance of the magnitude and gradient distributions }between vision and language models. \textbf{(c): The skewed distribution of the layer-wise scores} of SparseGPT.}
     \label{fig: dist}
     \vspace{-0.15in}
\end{figure}

However, most layer-wise pruning approaches focus on removing unnecessary weights or parameters from single-modal models, such as vision-only~\citep{chen2018shallowing,lazarevich2021post,yu2023x} or language-only~\citep{frantar2023sparsegpt,sun2023simple} architectures, which face significant limitations when extended to multimodal (e.g., vision-language) models due to the critical gap in weight/gradient distributions between different modalities, as shown in \Cref{fig: dist mag,fig: dist grad}. 
Unlike global pruning methods estimating the importance of each parameter based on the objective loss, layer-wise pruning without the guidance of global weight correlations suffers from obtaining relative significance across weights in different layers/modules due to such severe distributional imbalances in the different modality modules. For example, in Wanda, inputs for different layers significantly vary in their scales~\citep{Ba2016LayerN,Ioffe2015BatchNA}, and in SparseGPT, the importance of the local loss functions for different layers is also incomparable (Please see the skewed distribution of the local score shown in \Cref{fig: dist_sparsegpt}). 
This problem makes layer-wise pruning approaches usually set the pruning ratio to be a fixed value for all layers, resulting in a suboptimal compression rate and performance. This is also evident in the results from \cite{Singh2020WoodFisherES} that the global WoodFisher pruner outperforms its layer-wise counterpart by a large margin, especially at high sparsity regimes.

\section{ECoFLaP: Efficient Coarse-to-Fine Layer-Wise Pruning} \label{sec: method}

As discussed above, deploying a layer-wise pruning approach for large vision-language models is challenging due to three primary issues; \textit{distributional/architectural disparity between different modalities}, \textit{significantly large model size}, and \textit{lack of global knowledge}. To overcome such critical challenges, we propose \textit{\textbf{E}fficient} \textit{\textbf{Co}arse-to-\textbf{F}ine} \textit{\textbf{La}yer-wise} \textit{\textbf{P}runing} (\textbf{\MethodName{}}), to efficiently compute an adaptive pruning ratio for each layer with global importance scores (\textit{Coarse}), and then accurately remove the parameters layer by layer with local significance (\textit{Fine}).

\begin{figure}
    \centering
    \vspace{-0.2in}
    \includegraphics[width=0.9\textwidth]{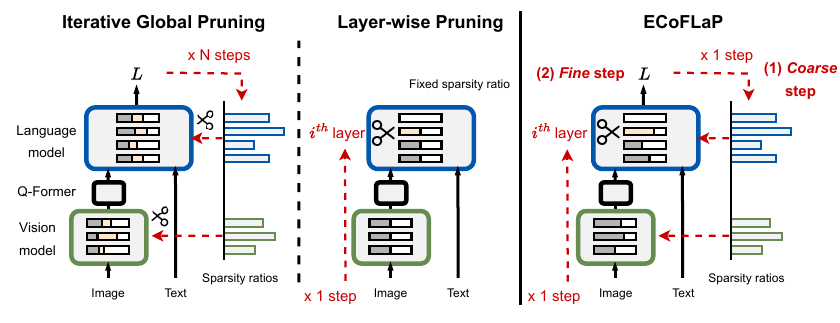}
    \vspace{-0.15in}
    \caption{\textbf{Illustration of our \MethodName{}} compared to \textit{global} and \textit{layer-wise pruning}. The boxes with a 
    \textcolor{fig_blue}{{\textbf{blue}}}, \textcolor{fig_green}{{\textbf{green}}}, and \textbf{black} 
    border denote the language, vision, and Q-Former modules, respectively. The dotted red arrows show the working flow of the algorithm. The \textcolor{fig_beige}{\textbf{beige}} color indicates the pruning of the current step (layer) is conditioned on the pruning decisions made in the preceding steps (layers), which are marked in \textcolor{fig_gray}{\textbf{gray}}. \MethodName{} first performs the efficient \textit{coarse} step to obtain the pruning ratio for each layer by leveraging the zeroth-order gradient, and then removes the uncritical weights in a layer-wise fashion in the \textit{fine} step.}
    \vspace{-0.1in}
    \label{fig: ecoflap}
    
\end{figure}

\subsection{The framework of coarse-to-fine layer-wise pruning}\label{subsec:ecoflap}
Our focus is to prune the BLIP-like multimodal architectures, as discussed in \Cref{ssec: framework of vl models}. Note that we compress the vision and language model in BLIP variants while keeping Q-Former intact because it is sufficiently lightweight, occupying only $\sim5\%$ of the parameters in the whole framework. Given the multimodal calibration data $\mathcal{D}=\{\textbf{x}^v_k, \textbf{x}^l_k\}^K_{k=1}$, a conventional layer-wise pruning method aims to find the sparse weight $\widehat{\textbf{W}}_i$ at layer $i$ via the corresponding local objective $\mathcal{L}_i$: 
\begin{align} \label{eq: layer-wise pruning}
    \widehat{\textbf{W}}_i = \arg \max \mathcal{S}(\textbf{W}_i | \widehat{\textbf{W}}_{i-1}, \mathcal{D}, \mathcal{L}_{i}), ~~~
    \text{s.t.}, \frac{|\widehat{\textbf{W}}_i|}{|\textbf{W}_i|} = ~p_i,
\end{align}
where $\mathcal{S}(\cdot)$ is the score function to compute the importance of the weight, $\widehat{\text{W}}_i$ is denoted as the pruned weight from $\text{W}_i$, and $p_i\%$ the desired sparsity at layer $i$. Basically, $p_i$ is the same for all layers in layer-wise pruning methods.

In our proposed Efficient Coarse-to-Fine Layer-wise Pruning (\MethodName{}) framework, we estimate the optimal sparsity for each layer in the vision-language model by computing the global importance of model weights. One straightforward direction to obtain the layer sparsities is to perform global pruning on the model for the target global sparsity $p$ by leveraging the global loss objective $\mathcal{L}$,
\begin{align}
    \widehat{\textbf{W}} = \arg \max  \mathcal{S}(\textbf{W} | \mathcal{D}, \mathcal{L}),~~~
    \text{s.t.}, ~\frac{|\widehat{\textbf{W}}|}{|\textbf{W}|} = p.
\end{align}
Then, we can estimate the layer sparsity via $p_i = |\widehat{\textbf{W}}_i| / |\textbf{W}_i|$.
However, this approach requires an expensive operation to obtain gradients for all weights in the large model and to extract the pruned weights via \texttt{argmax}, consuming enormous memory and computational resources.
Therefore, we introduce an alternative numerical approach, which computes the keep ratio (1 - sparsity ratio) based on importance scores linearly, to avoid expensive operations that can estimate the global importance score efficiently. First, we obtain the importance score for each weight via the global objective function, namely, $\textbf{s} = \mathcal{S}(\textbf{W} | \mathcal{D}, \mathcal{L})$, and then convert the scores to sparsity by three steps: (1) Find the total parameters that need to be selected based on $p$, (2) Normalize the scores, (3) Obtain the sparsity for each layer based on the number of parameters to be picked and the parameters of this layer. The example to determine the sparsity ratio for the $i^{th}$ layer of the model is shown below,
\begin{align} \label{eq: scores to ratios}
    \text{normalize}(\textbf{s}_i, \textbf{s}) &= \frac{\textbf{s}_i}{\sum{\textbf{s}}}, \\
    {p}_i = 1 - \frac{(\text{normalize}(\textbf{s}_i, \textbf{s}) \cdot N_{\text{select}})}{|\textbf{W}_i|}, &~~~~~\text{where}~ N_{\text{select}} = (1 - p) \cdot |\textbf{W}|.
\end{align}

We then inject the derived sparsity ratios back to \Cref{eq: layer-wise pruning} to prune the model layer-wisely. We also introduce a hyperparameter $p_{max}$ to control the maximum sparsity for each layer to avoid the layer collapse, which is discovered by previous literature~\citep{Tanaka2020PruningNN} that the global pruning approach may remove all the parameters for one layer.
To incorporate this hyperparameter in the \Cref{eq: scores to ratios}, we simply pre-pick the parameters for each layer to satisfy the max sparsity condition, subtracting the $N_{select}$ with the number of pre-picked parameters, and start the algorithm.
In our experiments, we find using the same sparsity ratio for the layers in one block (such as the feed-forward layer and attention layer in a transformer block) makes the results more robust, so we also extend the algorithm to determine group sparsity ratios by via group scores (aggregating all the layer scores in a block). 
For the layer-wise pruning step, we utilize Wanda, which combines the input activation of the layer with the weight magnitude as the local importance score, to remove uncritical parameters based on the obtained layer sparsity ratios.

\subsection{Zeroth-order global information}\label{subsec:ffpruning}

\begin{wrapfigure}[11]{r}{0.5\textwidth}
    \vspace{-1.0cm}
  \begin{center}
    \includegraphics[width=0.55\textwidth]{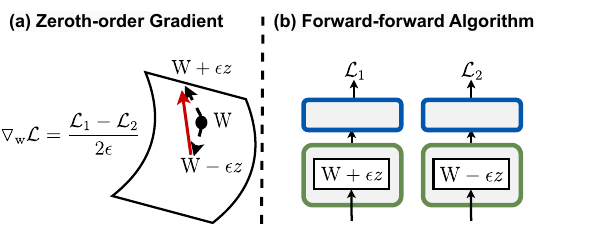}
  \end{center}
  \vspace{-0.1in}
  \caption{Illustration of (a) the zeroth-order gradient in weight space, and (b) the forward-forward algorithm.}  
  \label{fig: forward_forward}  
\end{wrapfigure}

In the previous section, we introduce our coarse-to-fine pruning framework to determine the pruning ratios for layer-wise pruning. Note that our goal is to prune models with billions of parameters, so we again focus on the one-shot setting in this step without any iterative refinement. To this end, our first attempt is to use the first-order backward gradient with weight magnitude~\citep{LeCun1989OptimalBD}, $|\textbf{W} | \cdot | {\triangledown_{\textbf{W}}\mathcal{L}(\textbf{W}, \mathcal{D})}|$, as the global importance score, and the performance improvement demonstrates the effectiveness of our framework (Please see \Cref{tab:main result} and \Cref{fig: acc_sparsity}). However, even though the first-order backward gradient is still tractable for most models. like less than 5B parameters, 7B LLaMA~\citep{Touvron2023LLaMAOA} requires more than \textbf{40GB} even with float16 weight type and batch size being one.

To further reduce the cost of computing gradients, we instead compute the zeroth-order approximated gradient by replacing the backpropagation with the forward-forward algorithm~\citep{Hinton2022TheFA,Malladi2023FineTuningLM}. 
We take the $i^{th}$ layer of a model as an example, the zeroth-order gradient of $\textbf{W}_i$ can be computed by perturbing the weight with Gaussian noises twice, that is,
\begin{align} \label{eq: zeroth-order grad}
    \norm{\triangledown_{\textbf{W}_{i}}\mathcal{L}(\textbf{W}_i, \mathcal{D})}_2 = \mathbb{E}_{d \sim \mathcal{D}}[ \mathbb{E}_{z \sim N(0, 1)} [ | \frac{\mathcal{L}(\textbf{W}_i + \epsilon z, d) - \mathcal{L}(\textbf{W}_i - \epsilon z, d)}{2\epsilon}| ]]
\end{align}
We hence use $\norm{\triangledown_{\textbf{W}_{i}}\mathcal{L}(\textbf{W}_i, \mathcal{D})}_2$ as the importance score of $\textbf{W}_{i}$. Note that we do not multiply it with the weight magnitude as solely using the approximated gradients works better empirically. The computation of \Cref{eq: zeroth-order grad} is extremely memory-efficient and the exact process for the $i^{th}$ layer is shown in the following: (1) First, we choose a random seed and then sample a noise $z$ (now we use 2x of memory of “this layer”), (2) we modify the model’s weight with $\textbf{W} + \epsilon z$ and compute the loss change $\mathcal{L}(\textbf{W}+ \epsilon z, d)$, (3) we then sample the same noise $z$ as the first step (by setting the same random seed), and we modify the model’s weight with $\textbf{W} - \epsilon z$, which can be done by subtracting $2 \epsilon z$ from the previous weight, and compute $\mathcal{L}(\textbf{W} - \epsilon z, d)$, (4) obtain the gradient norm by ($\mathcal{L}(\textbf{W} + \epsilon z) - \mathcal{L}(\textbf{W} - \epsilon z)) / (2\epsilon)$ (all terms are scalars). From the above process, for an N-layer network, we only need (N+1) times of one-layer memory, which is significantly memory-efficient than storing the whole gradients that cause 2N times of one-layer memory. This explains why our approach uses much less memory than first-order based methods, which also need to cache input activations. One may suspect we need to sample many noises to capture the loss of the landscape and to estimate the gradient accurately. However, our experiments and the previous studies~\citep{Malladi2023FineTuningLM} show that one noise is sufficient. We describe our approach in detail in \Cref{alg: ecoflap}.

\section{Experimental Setup} \label{sec: experimental setup}

We report evaluation metrics and more experimental details in \Cref{sec: experimental details}.

\noindent \textbf{Architectures.} We use multiple uni- and multi-modal architectures for experiments: the encoder-decoder vision of BLIP-2~\citep{Li2023BLIP2BL}, composed of pre-trained EVA-ViT (ViT-g/14 from EVA-CLIP)~\citep{Sun2023EVACLIPIT} and FlanT5~\citep{Chung2022ScalingIL}, is used for most experiments and ablation studies. And, we also extend our approach to BLIP~\citep{li2022blip} with ViT~\citep{dosovitskiy2020image} and BERT~\citep{Devlin2019BERTPO} backbones. In addition, we evaluate our approach solely on unimodal vision and NLP tasks with EVA-ViT, FlanT5, and LLaMA backbones. 

\noindent \textbf{Evaluation Datasets.} We evaluate the zero-shot ability of BLIP-2 on various datasets after pruning, such as VQAv2, OK-VQA, and GQA for visual question answering, NoCaps for image captioning, and Flickr30k for image-text retrieval. For BLIP, we evaluate the performance change of the BLIP fine-tuned on NLVR$^2$ and COCO captions. For unimodal models, we evaluate FlanT5 on MMLU, evaluate LLaMA 7B with WikiText, and evaluate EVA-ViT with ImageNet-1k.

\noindent \textbf{Baselines.} We compare \MethodName{} to several global pruning and layer-wise pruning approaches. For \textbf{Global Magnitude Pruning}, we perform global pruning based on the magnitude of the weight. \textbf{Gradient-based Pruning} is an iterative global pruning approach, where we use the multiplication of the first-order gradient and the weight magnitude as the importance score and prune the model to target sparsity in 3 iterations. \textbf{SparseGPT} is a layer-wise Hessian-based method. \textbf{Wanda} is also a layer-wise approach utilizing the multiplication of the weight magnitude and the norm of input activation as the local importance score. The comparison of the above pruning methods and their importance metric can be found in \Cref{tab: category}. We also compare our method with \textbf{UPop}, which prunes and re-trains the vision-language model simultaneously in a unified progressive pruning manner.

\section{Results} \label{sec: experiments}

\begin{table*}
    \centering
    \caption{Comparison of existing pruning approaches with \MethodName{} on the zero-shot performance with BLIP-2 at 0.5 sparsity. We report accuracy for visual question answering, CIDEr and SPICE for image captioning, and TR@1 (text recall) and IR@1 (image recall) for image retrieval. We also compute the macro average score over tasks and the GPU memory usage for each method. 
    }
    \label{tab:main result}
    \resizebox{0.98\textwidth}{!}{
    \begin{tabular}{lcccccccccc}
    \toprule

    \multirowcell{3}{Method} & \multirowcell{3}{Sparsity} & \multirowcell{3}{Method's \\ Mem. Usage \\ (GB)} & \multicolumn{3}{c}{Visual Question Answering}  & \multicolumn{2}{c}{Image Captioning} & \multicolumn{2}{c}{Image retrieval} & \multirowcell{3}{Macro \\ Avg.} \\
    \cmidrule{4-10}
    & & & VQAv2 & OK-VQA & GQA & \multicolumn{2}{c}{NoCaps} & \multicolumn{2}{c}{Flickr30k} &  \\   
    & & & \multicolumn{3}{c}{Accuracy} & CIDEr & SPICE & TR@1 & IR@1 & \\   
    \midrule
    Full Model  & 0\% & - & 63.1 & 41.1 & 44.1 & 121.7 & 15.8 & 97.6 & 89.7 & 62.1 \\
    \midrule
    Global Magnitude Pruning  & 50\% & 8.46 & 0.0 & 0.0 & 0.0 & 0.0 & 0.0 & 0.2 & 0.1 & 0.0 \\
    Gradient-based Pruning & 50\% & 22.4 & 56.9 & 35.3 & 41.7 & 96.8 & 13.5 & 93.0 & 82.4 & 55.4 \\
    SparseGPT & 50\% & 9.04 & 57.4 & 36.1 & 40.8 & 108.1 & 14.1 & \textbf{96.4} & \textbf{86.3} & 57.4 \\
    Wanda & 50\% & 8.87 & 57.9 & 35.3 & 42.2 & 106.9 & 14.1 & 95.1 & 84.6 & 57.2 \\
    \midrule
    \MethodName{} \\
    \cellcolor{gg}+ First-order Gradient & \cellcolor{gg}50\% & \cellcolor{gg}22.4 & \cellcolor{gg}\textbf{59.4} & \cellcolor{gg}\textbf{36.3} & \cellcolor{gg}42.3 & \cellcolor{gg}109.9 & \cellcolor{gg}14.2 & \cellcolor{gg}95.9 & \cellcolor{gg}86.2 & \cellcolor{gg}\textbf{58.2} \\
    \cellcolor{gg}+ Zeroth-order Gradient & \cellcolor{gg}50\% & \cellcolor{gg}8.93 & \cellcolor{gg}58.4 & \cellcolor{gg}35.6 & \cellcolor{gg}\textbf{42.9} & \cellcolor{gg}\textbf{110.2} & \cellcolor{gg}\textbf{14.3} & \cellcolor{gg}95.5 & \cellcolor{gg}85.8 & \cellcolor{gg}58.0 \\
    \bottomrule
    \end{tabular}
    }
\end{table*}
\begin{figure}[t]
    \centering
    \begin{subfigure}{0.29\textwidth}
        \centering
        \includegraphics[width=\textwidth]{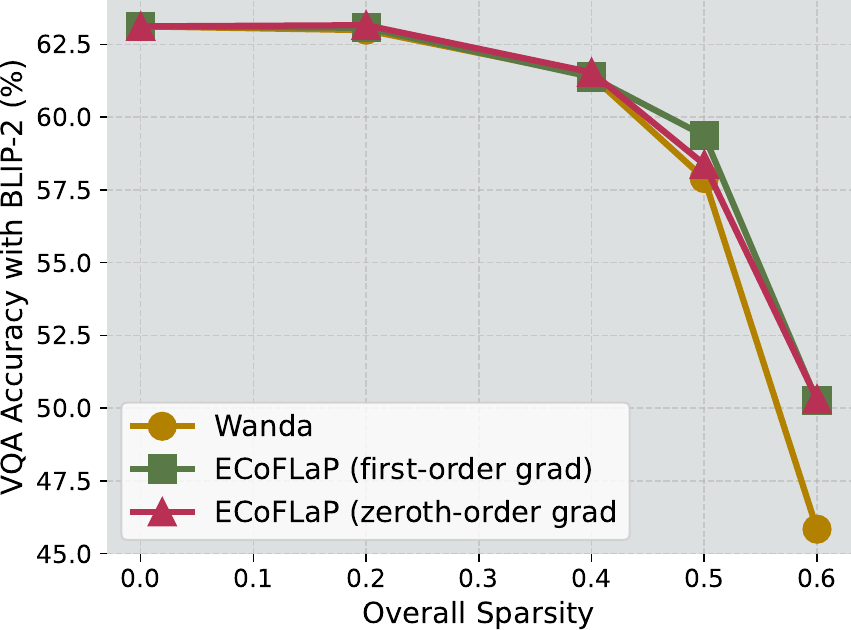}
        \caption{BLIP-2 on VQA}
        \label{fig: blip2_vqa}
    \end{subfigure}
    \hspace{0.15in}
    \begin{subfigure}{0.29\textwidth}
        \centering
        \includegraphics[width=\textwidth]{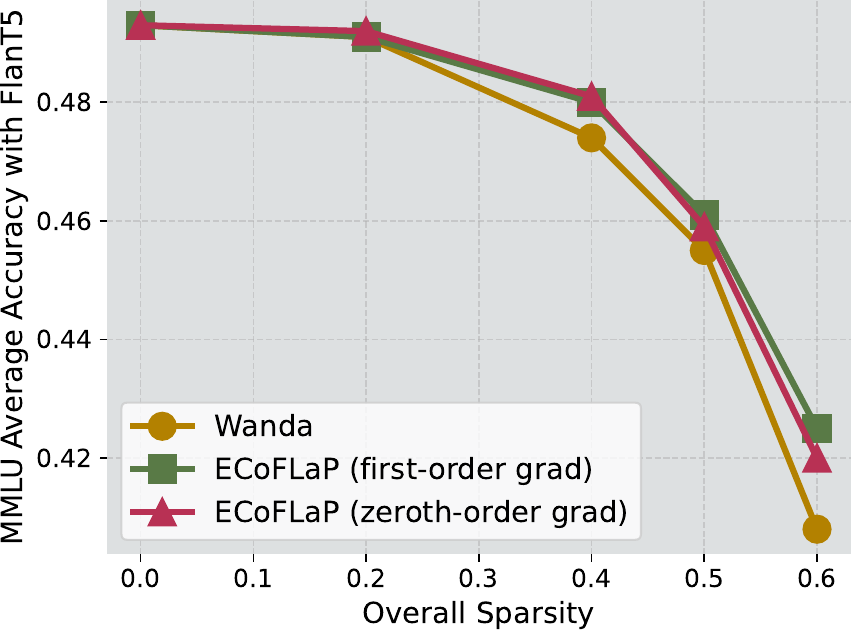}
        \caption{FlanT5 on MMLU}
        \label{fig: flant5_mmlu}
    \end{subfigure}
    \hspace{0.15in}
    \begin{subfigure}{0.29\textwidth}
        \centering
        \includegraphics[width=\textwidth]{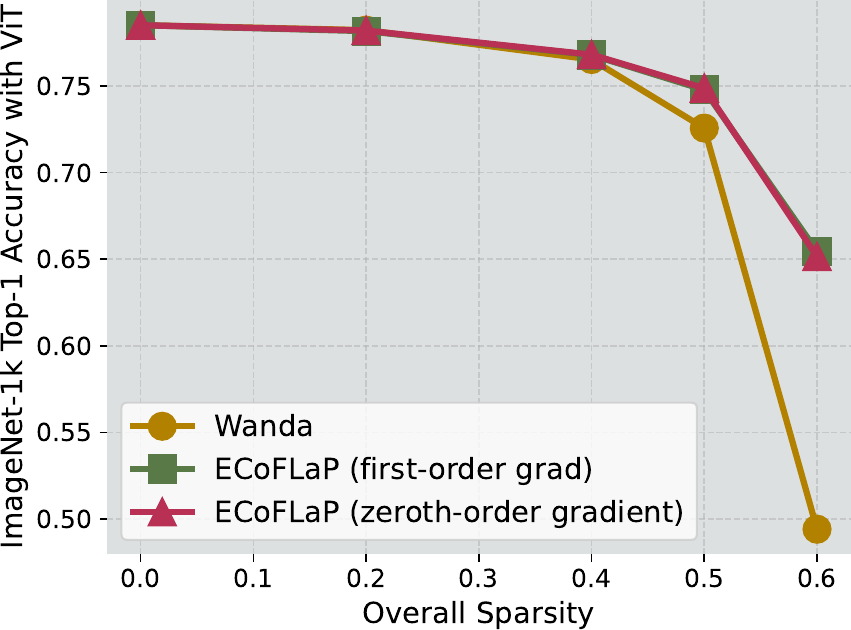}
        \caption{EVA-ViT on ImageNet}
        \label{fig: vit_imagenet}
    \end{subfigure}
    \caption{The accuracy and model sparsity trade-off for Wanda, \MethodName{} with first-order and zeroth-order gradient on both BLIP-2 (multimodal), FlanT5 (unimodal), and EVA-ViT (unimodal).}
    \label{fig: acc_sparsity}
\end{figure}

\subsection{Main experimental results} \label{ssec: main exps}

We demonstrate the zero-shot performance on various datasets using BLIP-2 pruned by our \MethodName{} and baselines at a 0.5 sparsity ratio in \Cref{tab:main result}. And we compare our approach with UPop using BLIP backbone on NLVR$^2$ and COCO captions under the fine-tuning and non-fine-tuning settings in \Cref{tab: upop and ecoflap}. We also provide the comparison of our approach and Wanda at various sparsities on both multi- and uni-modal models in \Cref{fig: acc_sparsity}. We summarize our observations as follows:

\noindent \textbf{\MethodName{} consistently outperforms Global Magnitude Pruning (GMP) and Gradient-based Pruning} as shown in \Cref{tab:main result}. GMP fails to generate meaningful results at 0.5 sparsity. In particular, we find that GMP suffers from pruning the vision module, EVA-ViT, so the degraded visual representation harms the multimodal model performance. Gradient-based Pruning performs significantly better by selecting crucial weights based on gradient information of the whole model, but it requires a larger memory than GMP due to the full pass to compute the gradients. On the other hand, compared to Gradient-based Pruning, our \MethodName{} with zeroth-order gradients uses only $40\%$ memory during pruning while improving average accuracy by $2.12\%$, $9.88\%$, and $3.41\%$, on visual question answering, image captioning, and image retrieval tasks, respectively.

\noindent \textbf{\MethodName{} outperforms SoTA layer-wise pruning baselines} since our \MethodName{} is able to exploit the global information. Our approach with zeroth-order gradient consistently achieves better performance against Wanda on all tasks and metrics by $1.1\%$, $2.3\%$, and $1.0\%$, on visual question answering, image captioning, and image retrieval tasks, respectively, showing the effectiveness of our coarse-to-fine framework. Furthermore, we highlight that the gap between Wanda and \MethodName{} becomes larger when the sparsity increases to 0.6, where \MethodName{} outperforms Wanda by 9.6\% on VQA (shown in \Cref{fig: blip2_vqa}). Our approach also surpasses SparseGPT by 1.4\% of relative improvement on the average score. Note that our zeroth-order \MethodName{} uses on-par memory budgets compared to layer-wise pruning methods, Wanda and SparseGPT, by avoiding expensive backpropagation computation. Additional results on the combination of \MethodName{} and SparseGPT are in \Cref{tab: sparsegpt}.

\begin{table}[t]
\centering
\caption{Zero-shot evaluation on 11 classification tasks with CLIP using $0.4$ of sparsity ratio. We additionally report results using local scores to compute layer sparsity ratios. $^{*}$FGVC, OFlowers, OPets, and SCars stand for FGVCAircraft, OxfordFlowers, OxfordPets, and StanfordCars, respectively.}
\label{tab: clip}

\resizebox{0.98\textwidth}{!}{
\begin{tabular}{lcccccccccccc}
\toprule
Methods      & Caltech101 & DTD & EuroSAT & FGVC$^{*}$ & Food101 & ImageNet & OFlowers$^{*}$ & OPets$^{*}$ & SCars$^{*}$ & SUN397 & UCF101 & Avg \\                       
\midrule
Full Model &  92.9 & 44.5 & 47.8 & 24.8 & 86.1 & 66.7 &    71.3  & 89.1 & 65.3 & 62.6 & 66.8 & 65.3 \\
\midrule
Wanda & 85.1 & 31.2 & 33.5 & 9.0 & 66.0 & 45.6 & 37.2 & 71.3 & 35.0 & 50.8 & 54.7 & 47.2 \\ 
SparseGPT & 90.5 & 41.5 & 47.4 & 19.2 & 79.4 & 50.8 & 59.0 & 86.4 & 53.4 & 55.3 & 63.2 & 58.7 \\
\midrule
\makecell[l]{Sparsities from \\ Local Scores} \\
w/ Wanda & 84.0 & 30.1 & 31.5 & 6.6 & 62.9 & 44.5 & 34.2 & 68.1 & 34.1 & 49.5 & 52.9 & 45.3 \\
w/ SparseGPT & 88.6 & 37.7 & 43.1 & 17.7 & 76.6 & 42.1 & 50.3 & 83.2 & 46.1 & 48.0 & 60.4 & 54.0 \\
\midrule
ECoFLaP \\
w/ Wanda & 87.5 & 37.9 & 43.8 & 16.0 & 72.8 & 53.5 & 57.0 & 80.3 & 46.9 & 58.2 & 62.4 & 56.0 \\
w/ SparseGPT & 90.0 & 42.4 & 45.4 & 22.5 & 81.2 & 56.5 & 64.1 & 88.3 & 56.8 & 60.2 & 63.4 & 61.0 \\
\bottomrule
\end{tabular}
}
\vspace{-0.2in}
\end{table}

\noindent \textbf{\MethodName{} surpasses baselines on pruning CLIP models by a significant margin}. 
Beyond pruning modularized VL models, \MethodName{} can also work well with joint VL models like CLIP. In \Cref{tab: clip}, we evaluate Wanda, SparseGPT, and our approach on CLIP by measuring the zero-shot accuracy on 11 classification tasks before and after pruning over the sparsity $0.4$, and our \MethodName{} excels against baselines while effectively mitigating the degeneration of zero-shot accuracy through the model pruning. 
In addition, we assess the impact of using global scores for computing layer sparsity ratios by comparing it with the approach using local scores derived from layer-wise pruning methods (namely, Wanda and SparseGPT). \textbf{The result shows that the strategy of using local scores leads to a decrease in performance compared to leveraging a uniform sparsity ratio}. This observation emphasizes the ineffectiveness of employing local scores for assessing layer importance (Please see our motivation in \Cref{fig: dist}).

\begin{wraptable}[12]{r}{0.43\textwidth}
    \centering
    \vspace{-0.2in}
    \caption{Performance comparison on BLIP at $0.5$ sparsity on NLVR$^2$ and COCO captions.}
    \label{tab: upop and ecoflap}
    \vspace{-0.1in}
    \resizebox{0.8\linewidth}{!}{
    \begin{tabular}{lcccc}
    \toprule    
    \multirowcell{2}{Method} & \multicolumn{2}{c}{NLVR$^{2}$} & \multicolumn{2}{c}{COCO cap.} \\
    & val & test & CIDEr & SPICE \\
    \midrule
    Full Model & 82.3 & 83.6 & 133.3 & 23.8  \\
    \midrule
    \multicolumn{5}{c}{\textit{w/o fine-tuning}} \\
    UPop & 76.9 & 77.8 & - & - \\
    Wanda & 78.3 & 78.1 & 97.1 & 18.4  \\
    \cellcolor{gg}\MethodName{} & \cellcolor{gg}\textbf{79.1} & \cellcolor{gg}\textbf{79.2} & \cellcolor{gg}\textbf{111.0} & \cellcolor{gg}\textbf{20.3} \\
    \midrule
    \multicolumn{5}{c}{\textit{w/ fine-tuning}} \\
    UPop & 80.3 & 81.1 & 128.9 & 23.3 \\
    \cellcolor{gg}\MethodName{} & \cellcolor{gg}\textbf{81.8} & \cellcolor{gg}\textbf{82.5} & \cellcolor{gg}\textbf{132.3} & \cellcolor{gg}\textbf{23.8} \\
    \bottomrule
    \end{tabular}
    }
\end{wraptable}

\noindent \textbf{\MethodName{} outperforms the multimodal model pruning baseline, UPop, in both non-fine-tuning and fine-tuning cases} as shown in \Cref{tab: upop and ecoflap}. Note that UPop requires re-training the backbone model in the pruning phase, which is significantly expensive in memory and computation, especially for large models. On the contrary, \MethodName{} finds an important subset of weights in the model in a single shot and then fine-tunes the pruned model. We also emphasize that our approach generalizes to different architectures and datasets, where BLIP uses a BERT-encoder for NLVR$^2$ (classification) and a BERT-decoder for COCO (image captioning).

\noindent \textbf{\MethodName{} also generalizes well to unimodal tasks/architectures.} In \Cref{fig: flant5_mmlu} and \Cref{fig: vit_imagenet}, we visualize the accuracy of MMLU with FlanT5 and ImageNet-1k with EVA-ViT  after pruning with various sparsity ratios (from $0.1$ to $0.6$). Since Wanda applies the same pruning ratio to all layers, performance degrades with increasing sparsity ($\sim40\%\uparrow$), which is catastrophic for some layers containing much more informative weights than others. Alternatively, the performance degradation caused by pruning weights can be significantly mitigated by proposing adaptive layer-wise pruning with dynamic sparsity for each layer based on estimated global weight importance. Furthermore, as shown in Table \ref{tab: llama}, our approach shows the ability to generalize to larger LLM, where we attain 10.6\% relative improvement over Wanda when evaluated on WikiText using perplexity with LLaMA 7B.

\subsection{Ablations and Visualization}

We provide ablation studies of our zeroth-order \MethodName{} on maximum sparsity for each layer $p_{M}$, the number of data $|\mathcal{D}|$, and the number of perturbed noises per sample. We report the zero-shot VQA accuracy obtained from the BLIP-T5 with 0.5 total sparsity. Note that we use the numerical method to obtain the sparsity ratios for experiments (Please see \Cref{eq: scores to ratios}). 

\noindent \textbf{Maximum Sparsity.} First, we observe that the maximum sparsity is critical as it affects the results considerably, and the optimal value happens at $0.6$, which is a conservative value given the target ratio is $0.5$. Similar to the findings discovered by~\cite{Tanaka2020PruningNN}, the algorithm might suffer layer collapse when the maximum sparsity is not set ($p_M = 1$). We find it simple but effective to introduce this hyperparameter as we would like to avoid using the iterative solution proposed by~\cite{Tanaka2020PruningNN} because we are targeting the one-shot pruning. Based on the result shown in \Cref{tab: max sparsity}, we simply set $p_M = p + 0.1$ throughout our experiments. 

\noindent \textbf{Number of Data and Noises.} In the next ablation experiment, we focus on understanding how many samples and noises we need to estimate the gradient with zeroth-order optimization. We first ablate the effect of the number of data by fixing the number of noises per sample to be one, and the result in \Cref{tab: num of samples} shows that our approach is robust to the number of samples across 16 to 128, and we thus using 32 samples for better efficiency. Next, we explore if having a more accurate gradient estimation by sampling more noises can improve the accuracy. In this ablation, we make sure to forward through the model the same number of times when using different numbers of noises, meaning that we cut the number of data to $1/n$ when we increase the number of noises to $n$. The result in \Cref{tab: num of noises} demonstrates that using more data samples is more effective compared to sampling more noises when the budget is fixed.

\begin{table}[t]
    \begin{minipage}{0.32\textwidth}
        
        \centering
        \caption{The ablation on the maximum sparsity.}
        \label{tab: max sparsity}
        \vspace{-0.1in}
        \resizebox{\textwidth}{!}{
        \begin{tabular}{lccccc}
        \toprule
    
        \multirowcell{2}{BLIP-2 \\ ($p=0.5$)} & \multicolumn{5}{c}{Max Sparsity} \\
        & 0.5 & 0.6 & 0.8 & 1.0 \\
        VQA & 57.9 & \textbf{58.4} & 57.1 & 51.6 \\
        
        \bottomrule
        \end{tabular}
        }
    \end{minipage}
    \hfill
    \begin{minipage}{0.32\textwidth}
        
        \centering
        \caption{The ablation on number of samples.}
        \label{tab: num of samples}
        \vspace{-0.1in}
        \resizebox{\textwidth}{!}{
        \begin{tabular}{lcccc}
        \toprule
    
        \multirowcell{2}{BLIP-2 \\ ($p=0.5$)} & \multicolumn{4}{c}{Number of Samples} \\
        & 16 & 32 & 64 & 128 \\
        VQA & 58.2 & \textbf{58.4} & 58.3 & \textbf{58.4} \\
        
        \bottomrule
        \end{tabular}
        }
    \end{minipage}
    \hfill
    \begin{minipage}{0.32\textwidth}
        
        \centering
        \caption{The ablation on number of noises (total  samples = 32).}
        \label{tab: num of noises}
        \vspace{-0.1in}
        \resizebox{\textwidth}{!}{
        \begin{tabular}{lcccc}
        \toprule
        \multirowcell{2}{BLIP-2 \\ ($p=0.5$)} & \multicolumn{4}{c}{Number of noises} \\
        & 1 & 4 & 8 & 32 \\
        VQA & \textbf{58.4} & 58.1 & 58.0 & 58.1 \\
        \bottomrule
        \end{tabular}
        }
    \end{minipage}
    \vspace{-0.1in}
\end{table}

\noindent \textbf{Visualization and Analysis.} We also provide the visualization of the sparsity ratios obtained by \MethodName{} (\Cref{fig: sparsity ratios}) and the loss landscape of BLIP-2 (\Cref{fig: loss landscape}) to justify the use of zeroth-order optimization. In \Cref{fig: sparsity ratios}, we observe that our \MethodName{} variants allocate sparsity ratios in a similar distribution, in favor of lower sparsity ratios for the visual model and higher sparsity ratios for the language model, and this alignment provides evidence that the zeroth-order gradient obtained by the forward-forward algorithm can be the accurate but cheaper alternative to the first-order gradient in importance estimation. The landscape shown in \Cref{fig: loss landscape} presents that BLIP-2 is located at a smooth basin with a single local minimum. This loss landscape justifies the success of utilizing zeroth-order optimization as we avoid sampling weights located in other basins that may cause an inaccurate gradient estimation. We also present a qualitative analysis of the pruned model in \Cref{sec: qualitative results}.

\section{Conclusion} \label{sec: conclusion}

This paper has investigated the challenges of pruning large-scale vision-language models due to the dilemma between immense memory/computational overhead from global pruning and the suboptimal model performance of layer-wise pruning. Existing global pruning methods require an expensive backpropagation process to compute the inverse Hessian to obtain global weight importance, which makes these approaches infeasible for current large LLMs and large LVLMs. On the other hand, layer-wise pruning approaches are highly efficient by pruning weights based on layer-wise local information, but often suffer from a lack of global importance, resulting in a significant performance degeneration. 
To address the limitations of these pruning approaches while enjoying their merits, we propose \MethodName{} that adaptively finds the optimal sparsity for each layer by efficiently estimating the global weight importance, and then accurately prunes the vision-language model in a layer-wise manner dependent on the obtained pruning ratios. We utilize the zeroth-order optimization to obtain the gradients with forward pass only and achieve superior performance over multiple uni- and multi-modal tasks against recent layer-wise pruning methods by a significant margin, while consuming only around 40\% of GPU memory compared to the iterative global pruning approach. We hope our new efficient pruning can help deploy quality yet compact vision-language models.
\section{Reproducibility Statement}
Our codes are based on the publicly available LAVIS~\citep{li-etal-2023-lavis}, Wanda~\citep{sun2023simple}, MMLU~\citep{Hendrycks2020MeasuringMM}, UPop~\citep{shi2023upop}, and CoOp~\citep{Zhou2021LearningTP}. The experimental setup and details can be found in \Cref{sec: experimental setup} and \Cref{sec: experimental details}, respectively.
We also have made our code publicly available.
\section*{Acknowledgments}
We thank Prateek Yadav for helpful discussions. This work was supported by ARO Award W911NF2110220, ONR Grant N00014-23-1-2356, and NSF-AI Engage Institute DRL-211263. The views, opinions, and/or findings contained in this article are those of the authors and not of the funding agency.

\bibliography{iclr2024_conference}
\bibliographystyle{iclr2024_conference}

\newpage
\appendix

\begin{algorithm}
\small
\caption{Efficient Coarse-to-Fine Layer-wise Pruning}\label{alg: ecoflap}
\KwIn{weights of the vision model $\textbf{W}^v$, weights of the language model $\textbf{W}^l$, calibration dataset $\mathcal{D}$, target sparsity $p$, maximum sparsity per layer $p_{max}$}
\KwOut{The pruned weights $\widehat{\textbf{W}}^v$ and $\widehat{\textbf{W}}^l$}

\stcp{\textit{Coarse} step: determine the sparsity ratios via global information}

$\textbf{s} \gets \mathcal{S}(\textbf{W}^v, \textbf{W}^l, \mathcal{D}, \mathcal{L})$ \stcp*{get the global importance score}

$\textbf{p} \gets \text{getSparityFromScores}(\textbf{s}, \textbf{W}^v, \textbf{W}^l, p, p_{max})$ \stcp*{obtain the sparsity via \Cref{eq: scores to ratios}}

\stcp{\textit{Fine} step: perform layer-wise pruning}

$\widehat{\textbf{W}}^v, \widehat{\textbf{W}}^l  = \{\}, \{\}$

\stcp{We initialize the $0^{th}$ weight to be the identity matrix, depicting the fact that the input of the first layer is the raw input.}

$\textbf{W}^v_0 \gets I$ 

\For{$i$ = $1$ to $M$}{
$\widehat{\textbf{W}}^v_i = \arg \max \mathcal{S}(\textbf{W}^v_i | \widehat{\textbf{W}}^v_{i-1}, \mathcal{D}, \mathcal{L}^v_{i}, \textbf{p}^{v}_i)$ \stcp*{prune vision weights layer by layer}

$\widehat{\textbf{W}}^v = \widehat{\textbf{W}}^v + \{\widehat{\textbf{W}}^v_i\}$
}

$\textbf{W}^l_0 \gets \textbf{W}^v_M$

\For{$i$ = $1$ to $L$}{
$\widehat{\textbf{W}}^l_i = \arg \max \mathcal{S}(\textbf{W}^l_i | \widehat{\textbf{W}}^l_{i-1}, \mathcal{D}, \mathcal{L}^l_{i}, \textbf{p}^{l}_i)$ \stcp*{prune language weights layer by layer}

$\widehat{\textbf{W}}^l = \widehat{\textbf{W}}^v + \{\widehat{\textbf{W}}^l_i\}$

}

\Return $\widehat{\textbf{W}}^v, \widehat{\textbf{W}}^l$

\end{algorithm}

\begin{figure}[h]

    \begin{minipage}{0.48\textwidth}
      \begin{center}
        \includegraphics[width=0.95\textwidth]{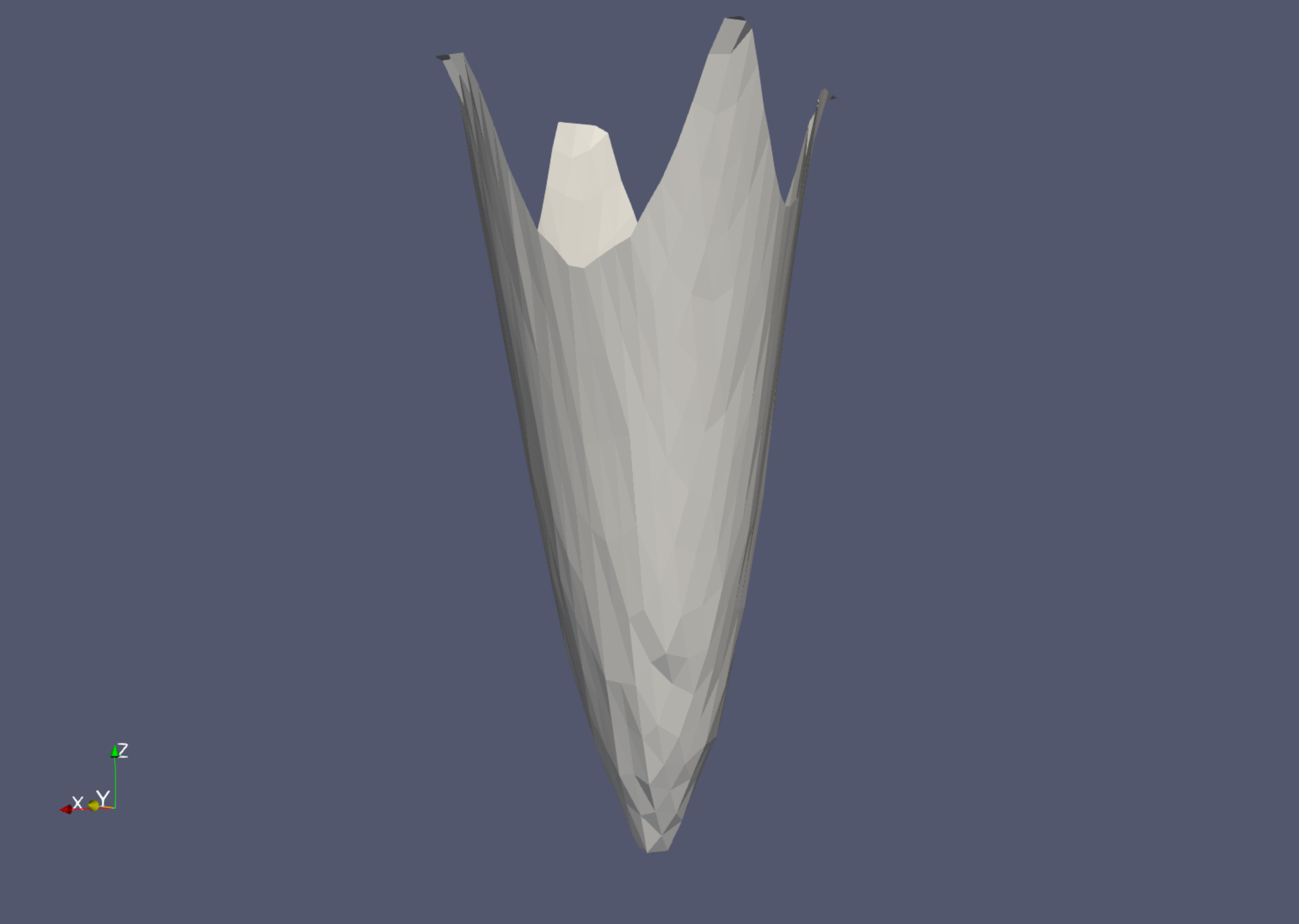}
      \end{center}
      \caption{The loss landscape of the BLIP-2 model.}
      \label{fig: loss landscape}
    
    \end{minipage}
    \hfill
    \begin{minipage}{0.48\textwidth}
        \centering
        \includegraphics[width=0.95\linewidth]{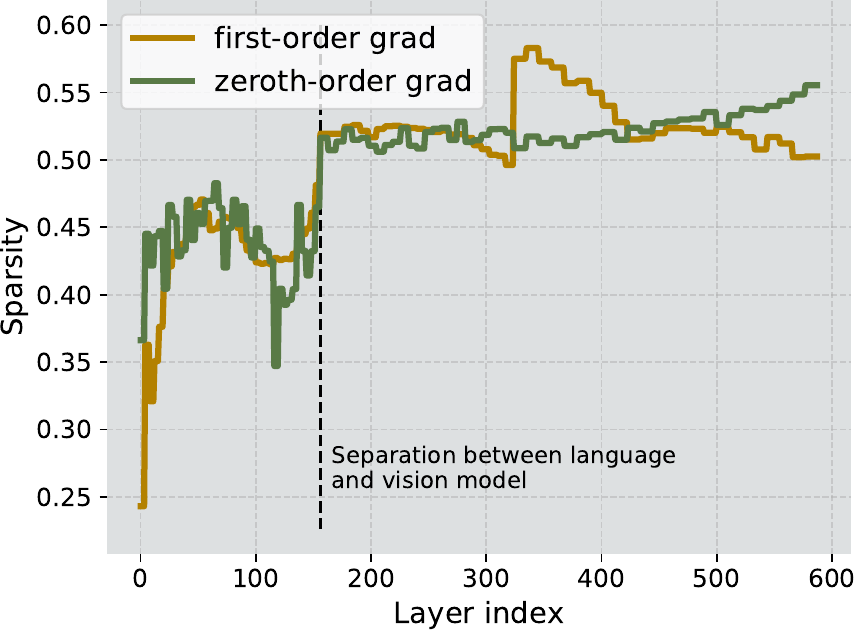}
        \captionof{figure}{Sparsity ratios obtained by \MethodName{}.}
        \label{fig: sparsity ratios}
    \end{minipage}
\end{figure}

\section{The loss landscape} \label{sec: loss landscape}

We present a visualization of the loss landscape~\citep{visualloss} for BLIP-2 in Figure \ref{fig: loss landscape}, where we achieve this by perturbing the pre-trained BLIP-2 model 2500 times to approximate the parameter space and using 256 samples from CC3M to compute the loss. Our visualization reveals a distinctive corn-shaped loss landscape, which is characterized by a single basin centered around the pre-trained weight of the BLIP-2. Moreover, the landscape is surprisingly smooth with a linear slope around the pre-trained weight. This smooth and corn-shaped loss landscape explains why we can estimate the gradient with zeroth-order optimization this successfully, as we avoid sampling weights situated in other basins that could lead to inaccurate gradient estimation.

\section{Results on LLaMA} \label{sec: llama results}

We further assess our approach's performance using perplexity on the WikiText dataset with LLaMA, a prominent open-source LLM. The results on the 7B model presented in Table \ref{tab: llama} demonstrate that \MethodName{} achieves a noteworthy 10.6\% relative improvement compared to Wanda when operating at 0.6 sparsity. Notably, in this experiment, we find that employing a larger $\epsilon$ value in Equation \ref{eq: zeroth-order grad} substantially enhances performance. This observation suggests that the zeroth-order estimation can benefit from exploring regions further from the target weight, a capability not inherent in gradients obtained through backpropagation.

\section{Experimental details} \label{sec: experimental details}

\noindent \textbf{Evaluation Datasets and Metrics.} We evaluate the zero-shot ability of BLIP-2 on various datasets after pruning. We use VQAv2~\citep{Goyal2016MakingTV}, OK-VQA~\citep{Marino2019OKVQAAV}, and GQA~\citep{Hudson2019GQAAN} for visual question answering, NoCaps~\citep{Agrawal2019nocapsNO} for image captioning, and Flickr30k~\citep{Plummer2015Flickr30kEC} for image-text retrieval. For BLIP, we evaluate the performance change of the BLIP, fine-tuned on NLVR$^2$~\citep{Suhr2018ACF} and COCO captions~\citep{Chen2015MicrosoftCC}. For unimodal models, we evaluate FlanT5 on 57 tasks in MMLU~\citep{Hendrycks2020MeasuringMM}, evaluate LLaMA 7B with WikiText~\citep{Merity2016PointerSM}, and evaluate EVA-ViT on image classification task with ImageNet-1k~\citep{Deng2009ImageNetAL}.
We report accuracy for question-answering datasets, NLVR$^2$, ImageNet-1k, and tasks in MMLU. Next, we use CIDEr and SPICE to evaluate image captioning tasks and use TR@1 (top-1 text recall) and IR@1 (top-1 image recall) for image retrieval tasks. Lastly, we report the model's perplexity of WikiText in evaluating the LLaMA. 

\noindent \textbf{Calibration Datasets.} Our approach utilizes a small subset (128 samples) of a dataset as the calibration data from CC3M~\citep{Sharma2018ConceptualCA}, ImageNet~\citep{Deng2009ImageNetAL} and C4~\citep{Raffel2019ExploringTL} for calibrating BLIP-2, unimodal EVA-ViT, and unimodal FlanT5 (and LLaMA). We use NLVR$^2$ and COCO captions as the calibration data for fine-tuning the BLIP backbone.

\noindent \textbf{Dataset splits.} For VQAv2, OK-VQA, and GQA, we use val, test, and test-dev split, respectively. We use the validation set for NoCaps and the test set for Flickr30k. In our BLIP experiments, we report results on both val and test set for NLVR$^2$, while we use the Karpathy test split~\citep{Karpathy2014DeepVA} for COCO captions. In our unimodal experiments, we use the publicly available test set for MMLU, and we use the validation set for ImageNet-1k. For WikiText, we report the perplexity on the validation set. 

\noindent \textbf{Datasets and Metrics for CLIP experiments.} We use a diverse set of 11 image classification datasets to evaluate the model's ability to recognize generic and specific objects: Caltech101~\citep{FeiFei2004LearningGV}, DTD~\citep{Cimpoi2013DescribingTI}, EuroSAT~\citep{Helber2017EuroSATAN}, FGVCAircraft~\citep{Maji2013FineGrainedVC}, Food101~\citep{Bossard2014Food101M}, ImageNet~\citep{Deng2009ImageNetAL}, OxfordFlowers~\citep{Nilsback2008AutomatedFC}, OxfordPets~\citep{Parkhi2012CatsAD}, StanfordCars~\citep{Krause20133DOR}, SUN397~\citep{Xiao2010SUNDL}, UCF101~\citep{Soomro2012UCF101AD}. We report accuracy for each dataset in \Cref{tab: clip}.

\noindent \textbf{Seed.} The default seed option for each codebase is used for evaluation: seed 42 for all the experiments for BLIP-2, FlanT5, and EVA-ViT, seed 0 for the LLaMA experiment, and seed 1 for the CLIP experiment.

\noindent \textbf{Hyperparameters.} For the coarse step, we set $p_M = p + 0.1$, $|\mathcal{D}| = 32$, and the number of noises to be 1 for \MethodName{} with zeroth-order gradient. $\epsilon$ is set to $1e^{-3}$ except for LLaMA, where we find $1e^{-1}$ works better, but we almost did not tune this hyperparameter. For \MethodName{} with first-order gradient, we use  $|\mathcal{D}| = 128$. 

\noindent \textbf{Resources.} All the experiments are done with one 40GB A100 or one 48GB A6000, except for ECoFLaP with first-order gradient on LLaMA we use 2x 48GB A6000.

\begin{table}
    \centering
    \caption{Evaluation of Wanda and \MethodName{} on LLaMA 7B at 0.6 sparsity by perplexity on WikiText.}
    \label{tab: llama}
    \resizebox{0.35\textwidth}{!}{
    \begin{tabular}{lc}
    \toprule

    LLaMA @ 0.6 sparsity & {Perplexity} \\
    \toprule
    Full Model & 7.26 \\
    Wanda & 10.68\\
    \midrule
    \MethodName{} \\
    + first-order gradient & 10.16 \\
    + zeroth-order gradient & 9.83 \\
    \bottomrule
    \end{tabular}
    }
    
\end{table}

\begin{table}[t]
\centering
\caption{The comparison of different methods in terms of their category and used importance measure.}
\label{tab: category}
\resizebox{1.0\textwidth}{!}{
\begin{tabular}{lccc}
\toprule
Methods      & Global or Layer-wise & Importance Measure \\                                                        \midrule
Global Magnitude Pruning  & Global &  $\| \textbf{W} \|_{ij}$ \\
Gradient-based Pruning  &  Global & $ \| \textbf{W} \| _{ij} \cdot \| \triangledown_{\textbf{W}_{i}}\mathcal{L} \| _{ij}$ \\
SparseGPT  & Layer-wise & $[\|\textbf{W}\|^ 2 / \text{diag} (\textbf{X}\textbf{X}^T + \lambda \textbf{I})^{-1}]_{ij}$   \\
Wanda   & Layer-wise & $\|\textbf{W}_{ij}\| \cdot \|\textbf{X}_{j}\|^2 $ \\
\makecell[l]{ECoFLaP \\ w/ Wanda and zeroth-order} & Mixed & \makecell{First Stage: $ \| \triangledown_{\textbf{W}_{i}}\mathcal{L} \| _{ij} $ \\ Second Stage: $\|\textbf{W} _{ij}\| \cdot \|\textbf{X} _{j}\|^2 $} \\
\bottomrule
\end{tabular}
}
\end{table}

\section{\MethodName{} on SparseGPT}

As the key contribution of our method for efficient estimation of layer-adaptive pruning rate for layer-wise pruning is orthogonal to SparseGPT, ECoFLaP can also be combined with SparseGPT. As shown in \Cref{tab: sparsegpt}, EcoFLaP (w/ SparseGPT) consistently achieves superior performance over multiple VL tasks.
 
Additionally, the performance improvement seems relatively smaller compared to ECoFLaP with Wanda (i.e., original EcoFLaP). We hypothesize that this is because the reweighting mechanism in SparseGPT is somewhat correlated to our dynamic sparsity ratios, where both of them are important to preserve the performance in high sparsity ratios.

\begin{table*}
    \centering
    \caption{Comparison of SparseGPT and \MethodName{} (with SparseGPT) on the zero-shot performance with BLIP-2 at 0.6 sparsity. We report accuracy for visual question answering, CIDEr and SPICE for image captioning, and TR@1 (text recall) and IR@1 (image recall) for image retrieval. We also compute the macro average score over tasks and the GPU memory usage for each method. 
    }
    \vspace{-0.1in}
    \label{tab: sparsegpt}
    \resizebox{1.0\textwidth}{!}{
    \begin{tabular}{lcccccccccc}
    \toprule

    \multirowcell{3}{Method} & \multirowcell{3}{Sparsity} & \multirowcell{3}{Method's \\ Mem. Usage \\ (GB)} & \multicolumn{3}{c}{Visual Question Answering}  & \multicolumn{2}{c}{Image Captioning} & \multicolumn{2}{c}{Image retrieval} & \multirowcell{3}{Macro \\ Avg.} \\
    \cmidrule{4-10}
    & & & VQAv2 & OK-VQA & GQA & \multicolumn{2}{c}{NoCaps} & \multicolumn{2}{c}{Flickr30k} &  \\   
    & & & \multicolumn{3}{c}{Accuracy} & CIDEr & SPICE & TR@1 & IR@1 & \\   
    \midrule
    SparseGPT & 60\% & 9.04 & 48.7 &  \textbf{27.9} & 35.1 & 101.0 & 13.4  &  95.2 & 85.0 & 51.8 \\
    \midrule
    \MethodName{} \\
    \cellcolor{gg} w/ SparseGPT & \cellcolor{gg}60\% & \cellcolor{gg}9.05 & \cellcolor{gg}\textbf{50.4} & \cellcolor{gg}\textbf{27.9} & \cellcolor{gg}\textbf{36.3} & \cellcolor{gg}\textbf{105.3} & \cellcolor{gg}\textbf{14.0} & \cellcolor{gg}\textbf{95.3} & \cellcolor{gg}\textbf{86.1} & \cellcolor{gg}\textbf{53.1} \\
    \bottomrule
    \end{tabular}
    }
    \vspace{-0.1in}
\end{table*}

\section{Qualitative comparison of the model before and after pruning} \label{sec: qualitative results}

We conducted an analysis comparing the predictions of the Full model (before pruning) and the model pruned by ECoFlaP (with 0.5 sparsity) on the GQA datasets, and presented three illustrative examples in \Cref{tab: qualitative examples}. We found one interesting observation: out of 12578 questions, the Full model correctly answered 1219 questions that our model did not, yet our model also accurately responded to 1072 questions where the Full model failed to respond (there are 4322 questions that both models answer correctly). This indicates that pruning does not always lead to performance degradation; in some instances, it may even enhance model robustness for certain questions. During our review of incorrect predictions by either our model or the Full model, we noticed that most were conceptually similar to the ground truth. This suggests that the pruned model still retains most of the capability. Furthermore, in some cases, such as with the 'horses running' example, both models provided answers that might be considered correct by many humans, even though they differ from the ground truth.

\begin{table}[t]
\centering
\caption{Qualitative examples to compare the predictions generated by the Full Model and the ECoFLaP-pruned model. We pick three example image-question pairs from GQA datasets. We observe the model after pruning does not always perform worse than the Full Model.}
\label{tab: qualitative examples}

\resizebox{\textwidth}{!}{
\begin{tabular}{lccc}
\toprule

\textbf{Images} & \begin{minipage}{.4\textwidth}
    \includegraphics[width=\textwidth]{figures/microwave_example.jpg}
\end{minipage} & \begin{minipage}{.4\textwidth}
    \includegraphics[width=\textwidth]{figures/car_example.jpg}
\end{minipage} & \begin{minipage}{.4\textwidth}
    \includegraphics[width=\textwidth]{figures/horse_example.jpg}
\end{minipage} \\
\midrule
\textbf{Question ID} & 202174017 & 20984557 & 20602949 \\
\midrule
\textbf{Questions} &  What shape is the microwave the stove is below? & Which kind of vehicle is parked on the street? & What is the horse running across? \\
\midrule
\textbf{Ground Truth} & rectangular & car & ground \\
\midrule
\makecell[l]{\textbf{Full Model's} \\ \textbf{Prediction}} & square & car & hill \\
\midrule
\makecell[l]{\textbf{ECoFLaP-pruned} \\ \textbf{Model's Prediction}} & rectangular & bicycle & rocks \\

\bottomrule
\end{tabular}
}
\end{table}

\end{document}